\documentclass[journal]{IEEEtran}
\ifCLASSINFOpdf
\else
\fi
\usepackage{graphicx} %插入图片的宏包
\usepackage{float} %设置图片浮动位置的宏包
\usepackage{subfigure} %插入多图时用子图显示的宏包
\usepackage{wrapfig}
\usepackage{stfloats}
\usepackage[justification=centering]{caption}
\usepackage{biblatex}
\usepackage{booktabs}  
\usepackage{multicol} 
\usepackage{color}%颜色模块

\usepackage{graphicx} % Required for inserting images
\usepackage{indentfirst}
\usepackage{amssymb}
\usepackage{amsmath}
\usepackage{hyperref}
\usepackage[figuresright]{rotating}

\begin{document}

\title{Unlocking the Potential: Multi-task Deep Learning for Spaceborne Quantitative Monitoring of Fugitive Methane Plumes}
%other titles
%Reducing Uncertainty of Methane Emission Monitoring in Landfills with Hyperspectral Satellite Images and Serial Deep Learning Networks
%Methane Plume Detection Algorithm Based on Weakly Supervised Multi-task Deep Learning


\author{Guoxin Si,
        Shiliang Fu,
        and Wei Yao \IEEEmembership{Senior Member,~IEEE}% <-this % stops a space
\thanks{G.Si, S.Fu and W.Yao are with Department of Land Surveying and Geo-Informatics, The Hong Kong Polytechnic University, Hong Kong e-mail: (wei.yao@ieee.org).}% <-this % stops a space
\thanks{Manuscript received April 19, 2005; revised September 17, 2014.}}

% that ends a line with a %, and do not let a space in before the next \thanks.

\markboth{Journal of \LaTeX\ Class Files,~Vol.~13, No.~9, September~2014}%
{Shell \MakeLowercase{\textit{et al.}}: Bare Demo of IEEEtran.cls for Journals}
% *** Note that you probably will NO,T want to include the author's ***
% You can use \ifCLANpeerreview for conditional compilation here if



\maketitle

\begin{abstract}

Methane ($CH_4$) is the second most prevalent greenhouse gas after carbon dioxide ($CO_2$), with a short atmospheric lifetime of approximately 9.14 years. Reducing methane emissions could quickly mitigate global warming. A methane plume is a localized concentration of methane. Methane emission monitoring involves three tasks: concentration inversion, plume segmentation, and emission rate estimation. The matched filter algorithm (e.g., Mag1c) for methane concentration inversion is sensitive to noise and reference signals. Research on methane plume segmentation is limited, often relying on subjective manual segmentation. Emission rate estimation typically uses the integrated mass enhancement (IME) algorithm, which is dependent on accurate meteorological measurements, limiting its applicability. Using PRISMA hyperspectral images from the WENT landfill in Hong Kong and the EnMAP hyperspectral dataset as a base map, a deep learning framework is proposed for quantitative methane emission monitoring from hyperspectral images based on physical simulation. We simulate methane plumes using large eddy simulation (LES) and create various fugitive methane concentration maps using the radiative transfer equation (RTE). We further augment the data to create a simulated EnMAP dataset $(https://github.com/Sigx093105/CH4_plume_dataset.git)$. We train a U-Net for concentration inversion, a Mask R-CNN for plume segmentation, and a ResNet-50 for emission rate estimation. These deep networks outperform conventional algorithms in validation accuracy and generalize well to RPISAM data. We combine tasks into multi-task learning models: MTL-01 and MTL-02, which outperform single-task models. This research demonstrates multi-task deep learning for quantitative methane monitoring and can be extended to various methane monitoring tasks.

\end{abstract}


% \fi
\IEEEpeerreviewmaketitle

\section{Introduction}
\IEEEPARstart{C}{u}rrently, landfills are still widely used around the world \cite{magazzino2020relationship}\cite{zhang2019greenhouse}. It is estimated that about 40-50\% of the greenhouse gases emitted by landfills are methane \cite{bogner2007waste}. Landfills account for approximately 5\% of global methane emissions and are considered one of the largest anthropogenic sources of $CH_4$ \cite{stocker2013technical}. Furthermore, their contribution to atmospheric methane has been increasing since the beginning of the 21st century \cite{canadell2021global}. The emission of greenhouse gases, including methane, not only affects the surrounding ecological environment \cite{iravanian2020types} but also has the potential to exacerbate extreme weather events in urban areas \cite{yang2021greenhouse}. From 1960 to 2019, the contribution of methane radiation stress accounted for 11\% of total radiation stress, making it the second largest greenhouse gas after carbon dioxide. Compared to other main greenhouse gases, such as carbon dioxide and nitrogen oxides, methane exhibits a relatively short atmospheric life of approximately 9.14 years \cite{canadell2021global}. This implies that reducing methane emissions may have a quicker impact on alleviating global warming compared to reducing emissions of other greenhouse gases.

To detect emitted methane plumes, airborne data (AVIRIS-NG) is commonly utilized due to small-scale methane emission sources \cite{foote2020fast}\cite{foote2021impact}. However, the advent of PRISMA, a new-generation hyperspectral satellite with a spatial resolution of 30 m, has made it feasible to retrieve spaceborne methane emissions \cite{cogliati2021prisma}. 
The monitoring process for methane plumes can be divided into three tasks: (1) methane concentration inversion, (2) methane plume segmentation, and (3) estimation of single plume flux rates. For methane concentration inversion, current methods predominantly rely on differential optical absorption spectroscopy (DOAS) and matched filter techniques. DOAS, although computationally intensive, is typically used for small-scale inversion in scenarios where emission point sources are known \cite{thorpe2017airborne}\cite{thorpe2013high}. On the other hand, matched filter algorithms are commonly applied for large-scale methane plume screening \cite{foote2021impact}\cite{thompson2015atmospheric}. However, it is important to note that the accuracy of the matched filter algorithm is heavily based on the precise modeling of the target spectrum and background estimation \cite{foote2021impact}\cite{theiler2006effect01}.

Few studies have been conducted on methane plume segmentation, but it is crucial to accurately delineate single plumes, as it significantly determines the accuracy of the estimation of the methane flux rate. Over the past few decades, a plethora of algorithms have been proposed for estimating point-source emission rates. However, most of these algorithms require local wind speed data as auxiliary information, which is difficult to obtain due to the limited availability of meteorological observation stations\cite{varon2018quantifying}
\cite{jongaramrungruang2022methanet}. addressed this limitation by employing simulated plumes and deep learning techniques to estimate emission rates without the need for wind speed information, although the presented network could only estimate the emission rate of single-source plumes. In practical scenarios, it is often necessary to segment individual plumes from an image and estimate the emission rates separately. 

Since the introduction of AlexNet in 2012, deep learning has witnessed rapid development and yielded remarkable results in various fields of remote sensing application\cite{JIA202314}\cite{JIANG202330}\cite{HE202476}. The U-net deep learning architecture was initially designed for semantic segmentation tasks\cite{ronneberger2015u}\cite{POLEWSKI2021297}, but also exhibits enormous potential for regression\cite{10.1145/3429309.3429328}. We regard methane concentration inversion as pixel-by-pixel regression, so U-net can be used to complete the task. Additionally, Mask R-CNN has proven to be highly effective in a wide range of instance segmentation tasks \cite{SANIMOHAMMED2022100024} and should also be applicable to Methane plume segmentation. Moreover, ResNet addresses the issue of gradient vanishing, enabling deeper networks to improve feature extraction capabilities\cite{he2016deep}\cite{shafiq2022deep}. Several earlier studies on emission rate estimation using shallow convolutional neural networks (CNN) have demonstrated feasibility, albeit with limited accuracy\cite{9027926}\cite{guanter2021mapping}. It is reasonable to apply ResNet to improve the emission rate estimation. And our goal is to leverage physically simulated plumes as training data for multi-task deep learning techniques to simultaneously perform methane concentration inversion, methane plume segmentation, and flux rate estimation in a unified approach.

In this article, we start by describing the foundational principles of the algorithms utilized in our study, including large-eddy simulation, radiation simulation, the methane plume concentration inversion algorithm, the methane plume segmentation algorithm (instance segmentation), and the plume emission rate estimation algorithm. Following this, we provide a detailed account of our experimental setup, which encompasses the research area, the collected data, and the procedures used to build our methane monitoring dataset. Lastly, we evaluate the performance of deep learning algorithms against traditional ones for the tasks of methane plume concentration inversion, methane plume segmentation (instance segmentation) and plume emission rate estimation. Furthermore, we present the two multi-task learning networks, MTL-01 and MTL-02, that we developed for these tasks. The major contributions of this paper are as follows:

1. A physics-informed deep learning scheme is proposed for spaceborne retrieval of methane emissions based on data simulation using Large Eddy Simulation (LES), Radiative Transfer Equation (RTE), and data augmentation techniques.

2. An instance segmentation algorithm is applied to solve the problem of isolating and identifying methane emission sources by plume segmentation.

3. We show that the serialization of multiple sub-tasks for methane emission monitoring leads to additional error accumulation, and designed a multi-task learning model to suppress the errors.



%\hfill September 17, 2023

%%%%%%%%%%%%%%%%Methods%%%%%%%%%%%%%%%%
\section{Method}

In this section, we present the principles of methane plume simulation, methane signal simulation, and methane concentration inversion. The two multi-task deep learning networks and relevant training mechanisms that we have designed are also presented.

\subsection{Simulation}

The simulation operation consists of two parts: simulation of methane plume emission and simulation of EnMAP-like satellite images with methane plume signal. For methane plume simulation, we utilized Large Eddy Simulation (LES), which is a computational fluid dynamics (CFD) method used to simulate turbulent phenomena in fluid flow\cite{sagaut2002large}. It combines the advantages of Direct Numerical Simulation (DNS) and Reynolds-averaged Navier-Stokes equations (RANS) methods, aiming to accurately predict the statistical characteristics of turbulent flow\cite{pope2001turbulent}\cite{domaradzki1997subgrid}. The core of LES lies in filtering the flow field into large- and small-scale vortices through a filtering operation. This can be achieved by applying a low-pass filter to remove high-frequency small-scale vortices while retaining the low-frequency large-scale vortices\cite{domaradzki1997subgrid}. The filtered flow field equations form the basis of LES. The governing equations for LES are based on Reynolds-averaged Navier-Stokes equations (RANS)\cite{germano1991dynamic}, which are solved to simulate motion and turbulent effects in the fluid \cite{10.1115/1.3124648}:

\begin{equation}\label{01}
\frac{\partial \bar{u}_i}{\partial t} + \frac{\partial}{\partial x_j}(\bar{u}_i \bar{u}_j) = -\frac{1}{\rho}\frac{\partial \bar{p}}{\partial x_i} + \nu \frac{\partial^2 \bar{u}_i}{\partial x_j \partial x_j} - \frac{\partial}{\partial x_j}(\bar{u}_i' u_j')
\end{equation}
where $\bar{u}_i$ is the time-average speed, $t$ is the temporal change in fluid motion, $x_i$ is the position of the fluid in a certain direction,  $\bar{p}$ is the average pressure of a fluid at a location, $\rho$  is fluid density, $v$ is kinematic viscosity\cite{chagovets2007effective}, and $\bar{u}_i' u_j'$ is the reynolds stress\cite{launder1975progress}.

For simulation of EnMAP remote sensing images with methane plumes, we merged synthetic plume signals with real EnMAP remote sensing data\cite{canadell2021global}. First, plumes of different shapes, sizes, and concentrations were generated by simulation and assigned corresponding spectral characteristics based on physical radiation properties. Then we fused the synthetic plumes with real EnMAP remote sensing images to generate simulated images with realistic background textures and synthetic methane spectrum. Such simulated images are crucial for training and evaluating our algorithms, as they can simulate real observation scenarios under controlled experimental conditions\cite{foote2020fast}:
\begin{equation}\label{02}
    L(\lambda){'}=L(\lambda)*T_{plume}
\end{equation}
where $L(\lambda)$ denotes the real EnMAP image background spectrum and methane signal, $L(\lambda){'}$ is the simulated spectrum, $T_{plume}$ is the simulated methane absorption cross-section. And $T_{plume}$ is generated by the summation of the radiative transfer equation (RTE), multiplied by the layered methane concentration and corresponding dry air density.

By combining these two steps, we can simulate methane plumes of various shapes, sizes, and concentrations and inject them into EnMAP data to generate simulated hyperspectral satellite images, which contain methane plume signals overlaid with a realistic scene background. This provides a solid foundation for our research and allows us to validate the performance and reliability of our methodology in real-world scenarios. 

\subsection{Methane Concentration Inversion}
Methane concentration inversion typically refers to a technique or process utilized to retrodict (or estimate) the concentration of methane on the earth's surface or in the atmosphere from satellite images or other remote sensing data. This process involves the application of mathematical models and algorithms to transform satellite imagery into quantitative assessments of methane concentrations.

Mag1c is a matched filter method that effectively aligns trace gas concentration path lengths by incorporating sparsity and reflectivity correction. Mag1c is also one of the most advanced matched filter methods currently available and provides a mature and operational software interface\cite{foote2021impact}. The calculation formula for Mag1c is as follows\cite{foote2020fast}:

\begin{equation}\label{03}
    a=\frac{(L_i-u)^{T}C^{-1}(t(u))}{(t(u))^{T}C^{-1}(t(u))}
\end{equation}
where $a$ is methane concentration, $L$ is radiation intensity, $t(u)$ is mean radiation, and $C$ is covariance matrix.

Methane concentration inversion in the West New Territories (WENT) region was performed using PRISMA data and the mag1c method. The results of this inversion were evaluated against a deep learning model to validate its generalization capabilities. Considering the intricate distribution of land and sea areas in Hong Kong, as well as the challenging rugged terrain and dense urban construction, it can be expected that the surface conditions in the region significantly influence the results of methane concentration inversion. To address this, we use a masking method based on K-means to run Mag1c\cite{funk2001clustering}. 

In the realm of deep learning technologies, U-net is a prevalent and easily implementable architecture, frequently used for image semantic segmentation tasks\cite{ronneberger2015u}\cite{POLEWSKI2021297}\cite{zang2021land}. Our rationale for selecting U-net as the model for methane concentration inversion comes from the following points: 1. We aim to achieve inversion of the methane concentration using a network that is as simplistic as possible. 2. Within the U-net framework, semantic segmentation is realized in a regression form, making it conveniently adaptable to the task of methane concentration inversion. 3. U-net can easily produce output images with the same length and width as the input image. Typically, the input and output channels of the U-net are of equal size. However, to accommodate the requirements for the inversion of methane concentration, an additional layer of $1*1$ convolution is added after the U-net encoding-decoding structure\cite{lin2013network}, and the loss function is replaced for the regression task. The loss functions used include the mean squared error (MSE) and $SmoothL_1$: 

\begin{equation}\label{04}
    loss_{mse}=\frac{1}{n} \sum_{i=1}^{n} (y_i - \hat{y_i})^2
\end{equation} 

\begin{equation}\label{05}
loss_{SmoothL1}= \begin{cases}  
\frac{1}{2} (y - \hat{y})^2 & \text{if } |y - \hat{y}| < 1 \\  
|y - \hat{y}| - \frac{1}{2} & \text{otherwise}  
\end{cases}  
\end{equation}
where $y$ is the true label, $\hat{y}$ is the predicted value, $n$ is the number of samples, and $|y - \hat{y}|$ represents the absolute difference between $y$ and $\hat{y}$. 

Before starting the training phase, various preprocessing techniques were applied to both labels and images. In particular, we computed the mean and standard deviation of all non-zero pixel values in the labels to achieve normalization. Referring to the pre-processing methods of the DOAS algorithm \cite{wagner2004max}\cite{platt2008differential}, the subsequent formula was utilized to preprocess the images: 
 \begin{equation}\label{06}
	{I_{after}=\frac{\log(I_{before})}{10}}
 \end{equation} 
 where $I_{before}$ denotes the image prior to preprocessing, and $I_{after}$ denotes the same image post preprocessing. Both EnMAP and PRISMA have pixel values within the range of 0-10 after taking the logarithm, and dividing these values by 10 normalizes all preprocessed pixel values to fall within the range of 0-1. This technique helps to generalize the model across various sensors. The network is optimized using two loss functions, MSE and SmoothL1, starting with an initial learning rate of 0.001. The learning rate is decreased by 75\% every 10 epochs, over a total of 200 epochs.

\subsection{Methane Plume Segmentation}
Instance segmentation is a classical task in computer vision, and methane plume segmentation is essentially an instance segmentation task. Previous research work on methane plume segmentation is limited, and many studies are based on subjective segmentation criteria\cite{gorrono2023understanding}. Some studies suggested using traditional algorithms for the segmentation of methane plumes, with contour tracking algorithms considered a good choice due to their efficiency and stability. On the other hand, there are many deep learning-based instance segmentation methods, among which Mask R-CNN is the most classical and widespread one\cite{he2017mask}. It is a two-step instance segmentation algorithm that has proven to achieve high performance in a large number of datasets for a wide range of applications. Although new instance segmentation algorithms are continuing to emerge, some of the algorithms sacrifice segmentation accuracy for real-time operation\cite{terven2023comprehensive}, which is not necessarily required for our case. Other algorithms may achieve higher segmentation accuracy\cite{wang2020solov2}, but require an increasingly complex optimization process during training, which is not cost-effective. Therefore, Mask R-CNN is chosen as the deep learning algorithm for methane plume segmentation and active contour algorithms as the traditional counterpart. Typically, Mask R-CNN takes three image channels as inputs, but we modified the input layer to allow flexible alteration between 1 or 41 input channels to accommodate different segmentation requirements\cite{foote2020fast}, since the methane distribution map has only one band and EnMAP contains 41 bands after excluding the water vapor and carbon dioxide sensitive bands. The original loss function of Mask R-CNN is used:
\begin{equation}\label{06}
 L = L_{cls} + L_{box} + L_{mask}
\end{equation}
where $L$ is the total loss function, $L_{cls}$ is the category loss function, which is based on the cross-entropy loss, $L_{box}$ is the bounding box loss function(As Mask R-CNN is a two-step instance segmentation model,its loss function inevitably includes the loss of object detection task), which is based on the $Smooth L_1$ loss, and $L_{mask}$ is the mask loss function, which typically uses the binary cross entropy:
\begin{equation}\label{07}
 L_{cls} = - \sum_{i=1}^{N} y_i \log(\hat{y}_i)
\end{equation}
where $y_i$ denotes the true label, $\hat{y}_i$ is the probability predicted by the model, and $N$ is the number of samples.
\begin{equation}\label{08}
 L_{box} = \sum_{i=1}^{n} \text{Smooth}_{L1}(x_i - \hat{x}_i)
\end{equation}
where $n$ is the number of bounding boxes considered in the current batch, $x_i$ represents a parameter set (including the center coordinates, width and height of the box) of the $i$-th predicted bounding box, $\hat{x}_i$ is the corresponding ground truth parameter for the $i$-th bounding box. During the training process, we use the minimum enclosing rectangle of the ground truth masks as the bounding box (for the predicted and ground truth bounding box). 

\begin{equation}\label{10}
\text{Smooth}_{L1}(x) =   
\begin{cases}   
0.5x^2 & \text{if } |x| < 1 \\  
|x| - 0.5 & \text{if } |x| \geq 1   
\end{cases}  
\end{equation}
where $x$ is the difference between the predicted value and the true value.

\begin{equation}\label{11}
 L_{mask} = - \text{IoU}(b, \hat{b}) + \frac{\rho^2(b, \hat{b})}{c^2}
\end{equation}
where $b$ and $\hat{b}$ respectively denote the ground truth binary mask and the predicted binary mask, $\text{IoU}(b, \hat{b})$ indicates the intersection over the union between these two binary masks, $\rho^2(b, \hat{b})$ is the Euclidean distance between the center points of two binary masks, and $c$ is a hyperparameter typically set to 10.

\subsection{Emission Rate Estimation}

There are two scenarios for estimating emission rates. The first scenario involves estimating the emission rates using a supervised deep model with training samples collected specifically for this purpose. The goal here is to evaluate the performance of the estimation algorithm under ideal conditions. The second scenario refers to the estimation of emission rates based on plume segmentation results (MTL-01), where the objective is to evaluate and improve the accuracy of plume instance segmentation.

The emission rate of a methane plume is commonly estimated using the integrated mass enhancement (IME) mode\cite{guanter2021mapping}. This mode involves the following calculations:

\begin{equation}\label{12}
    Q= \frac{U_{eff}\cdot IME}{L}
\end{equation}
where $U_{eff}$ is the effective wind speed in $m/s$, $L$ is the plume length $m$, $IME$ refers to the integrated mass enhancement. And the calculations for $U_{eff}$, $L$, and $IME$ are as follows\cite{varon2018quantifying}\cite{guanter2021mapping}:
\begin{equation}\label{13}
    U_{eff}=0.34 \cdot U_{10} + 0.44
\end{equation}
where $U_{10}$ is the wind speed at 10 meters.
\begin{equation}\label{14}
    L=\sqrt{A_M}
\end{equation}
where $\sqrt{A_M}$ is the area of a plume.
\begin{equation}\label{15}
    IME=k\sum_{i=1}^{n_p}\hat{\alpha}(i)
\end{equation}
where $n_p$ is the number of pixels in a plume, $\hat{\alpha}(i)$ is the value of $ith$ pixel, and $k$ is a scale factor equal to $5.155 \cdot 10^{-3} kg \slash ppb$.

ResNet is a deep learning method that uses residual blocks, which enables the neural network to learn the residual map by effectively avoiding issues such as gradient vanishing and explosion\cite{he2016deep}. ResNet is widely applied in various application fields, including image recognition, object detection, and speech processing, yielding impressive results\cite{veit2016residual}\cite{shafiq2022deep}. However, in this study, we pay particular attention to the application of ResNet in regression tasks, specifically for the complex and important task of emission rate estimation.
Since the emission process is often affected by a variety of environmental factors and source dynamics, this makes emission rate estimation a highly nonlinear and complex problem. To meet this challenge, we chose ResNet-50 as our base network architecture for the following reasons: 1. Powerful feature extraction capabilities: ResNet-50 is able to learn and extract rich hierarchical features in an image through deep structure and residual connection, which is essential to capture subtle changes in methane emission from images. 2. Suitable for regression tasks: although ResNet-50 was originally designed for classification tasks, it can be easily adapted to regression tasks, such as emission rate estimation, by adjusting the output layer of the network (replacing the softmax layer with a linear layer).

We trained the ResNet-50 network with samples to perform the flux-rate estimation task. Subsequently, ResNet-50 was used as a pretrained network to validate the rationale of the plume segmentation results and assist in optimizing the instance segmentation network.

\subsection{Multi-task Learning I (MTL-01)}

Experimental results indicated that the sequential prediction process involving Mask R-CNN followed by ResNet-50 led to substantial errors in emission rate estimation due to Mask R-CNN's false positives and false negatives (Fig. \ref{fig:picture009}). While these errors could be tolerable in the context of plume segmentation, we have developed a multi-task network named MTL-01 to tackle this issue. This network concurrently trains both Mask R-CNN and ResNet-50.

\begin{figure*}[hb]%H为当前位置，!htb为忽略美学标准，htbp为浮动图形
\centering %图片居中
\includegraphics[width=1\textwidth]{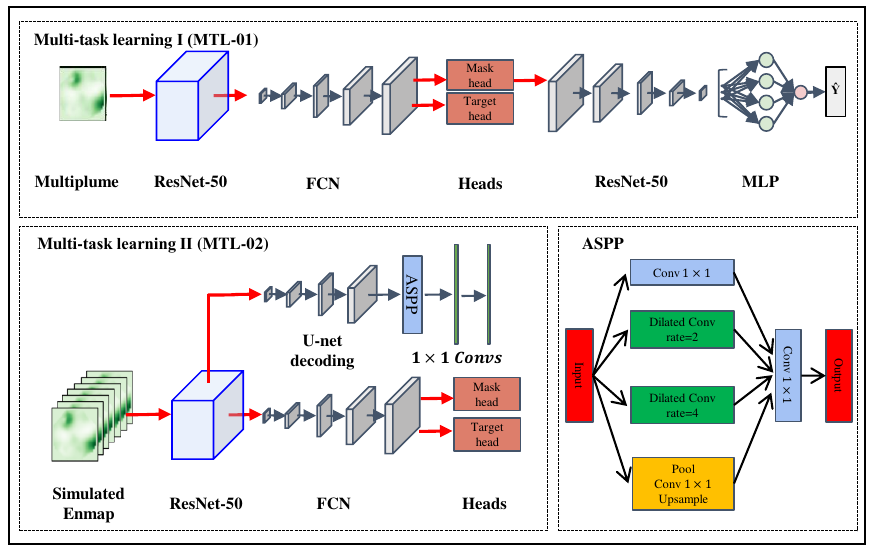} 
\caption{The structure of the multi-task learning model}
  \small\centering
  The first multi-task learning model (MTL-01) is a sequential architecture combining a Mask R-CNN for plume segmentation and a ResNet-50 for estimating plume emission rates. The segmented plumes from Mask R-CNN are fed into ResNet-50 in real-time for emission rate loss calculation. The total loss for MTL-01 is the weighted sum of the individual losses of both networks. 'Multiplume' describes a methane concentration map containing multiple plumes and noise. ResNet-50 is a 50-layer deep residual network. A Fully Convolutional Network (FCN) integrates semantic information from both low-level and high-level images via skip connections. In Mask R-CNN, 'heads' are responsible for detecting objects and extracting masks. MLP stands for multi-layer perceptron. The second multi-task learning model (MTL-02) detects methane concentration and plume boundaries simultaneously. This model consists of a Mask R-CNN for plume segmentation and a U-Net for methane concentration inversion, both sharing a ResNet-50 feature extractor. While the Mask R-CNN’s decoding structure is unchanged, the U-Net decoding structure includes four skip connections, an ASPP module, and two 1x1 convolutional layers. 'Simulated EnMAP' refers to EnMAP data augmented with simulated methane plume signals. The ASPP module comprises a 1x1 convolutional layer for channel reduction and multiple parallel atrous convolutional layers with adjustable dilation rates to capture features at different scales. 'Conv 1x1' indicates a 1x1 convolution layer, 'Dilated Conv' denotes an atrous convolution layer, and 'Pool' refers to a pooling layer.
  \label{fig:picture001}
\end{figure*}

Considering that the loss function of Mask R-CNN measures instance segmentation and does not consider error propagation, we design a serial network (MTL-01), which takes the segmentation results of Mask R-CNN as input to ResNet-50 and calculates additional loss terms for back-propagation to assist in optimizing Mask R-CNN (Fig. \ref{fig:picture001}). The ResNet-50 in the serial network is a pre-trained model, whose parameters are not updated. During the training process, the erroneous (over-)segmentation masks from Mask R-CNN are mainly categorized into two types. The first type is small false positive patches escaping from NMS(Non-Maximum Suppression) filtering\cite{he2017mask}, which were only found to slightly contribute to errors in subsequent emission rate estimation rather than main error source of the serial network. Simulation experiments indicate that the IME value of plumes is almost never below 300. Hence, we disregard false positive patches with IME values below 300 pixels, addressing this type of error using an object detection loss term. The second type of error consists of large false positive patches, often wrongly separated from overlapping plumes, causing substantial inaccuracies in emission rate estimation. To mitigate this, we introduce an additional loss term, $loss_{ER}$, for MTL-01. Although $loss_{ER}$ is still based on smoothL1, segmentation results with IME values below 300 pixels are excluded from its calculation. Missed masks are treated as having an emission rate estimate of 0 kg/h, and the loss is computed accordingly:

\begin{equation}\label{16}
loss_{ER}= \begin{cases}  
Smooth_{L1}(y_i , \hat{y_i}) & \text{if } y_i\in TP \\  
Smooth_{L1}(0 , \hat{y_i}) & \text{if } y_i\in FP
\end{cases}  
\end{equation}

 In MTL-01, we consider the $loss_{ER}$ term as a corrected loss term and assign it a coefficient lambda, while the original loss function of Mask R-CNN remains unchanged. Therefore, the loss function of MTL-01 is as follows:
\begin{equation}\label{17}
    loss_{MTL\--{01}}=loss_{Mask R\,\--{CNN}} + \lambda\cdot loss_{ER}
\end{equation}

In this equation, $loss_{Mask R\,\--{CNN}}$ denotes the mask R-CNN loss function, $loss_{ER}$ denotes the loss function of emission rate estimation, and $\lambda$ is the coefficient. Based on our experiments, we found that $\lambda = 0.1$ is optimal.

%-------------------Fig2
%-------------------Fig2

\subsection{Multi-task Learning II (MTL-02)}

The process of methane concentration inversion entails estimating pixel-specific values using hyperspectral images, whereas methane plume segmentation is concerned with identifying plume masks within the same sets of images. Given their structural similarity and logical relationship, using multi-task learning models might improve the accuracy of both methane concentration inversion and plume segmentation tasks. Hence, a multi-task learning framework that merges U-net and Mask R-CNN architectures has been devised.

In MTL-02 (Fig.\ref{fig:picture001}), the mask R-CNN network architecture remains unchanged, and the U-net decoding structure is connected to the mask R-CNN backbone network. Considering the output sizes of the stages in the ResNet-50 backbone network, we added an additional output to the first convolution block of ResNet-50 to ensure that the size of the U-net decoding result matches the input size, similar to LinkNet\cite{zhou2018d}. It should be noted that the deepest output of ResNet-50 is not involved in the construction of the U-net decoding structure but only participates in the FCN\cite{long2015fully} structure of Mask R-CNN. To address the potential underfitting of the U-net decoding structure due to the max-pooling layer situated between two outputs of ResNet-50, we introduced an Atrous Spatial Pyramid Pooling (ASPP) layer directly after the U-net decoder. This ASPP layer consists of two dilated convolutions with dilation rates of 2 and 4. ASPP\cite{chen2017deeplab}, which builds on dilated convolutions and SPP (Spatial Pyramid Pooling)\cite{he2015spatial}, addresses the challenge of balancing the reduction in feature map resolution with the need to capture a large receptive field for image feature extraction. ASPP leverages dilated convolutions with varying dilation rates to encapsulate the image context at multiple scales. In this approach, the input feature map is subjected to several dilated convolutions with different rates, which are then combined. To further expand the model's receptive field, the resulting feature maps from these dilated convolutions are subjected to average pooling.

In the loss function for $MTL-02$, a traditional approach is employed to use a weighted average of the individual loss functions for multiple tasks\cite{bruggemann2021exploring}:

\begin{equation}\label{18}
    loss_{MTL\--{02}}=w_1\cdot loss_{U\--{net}}+w_2 \cdot loss_{Mask R\,\--{CNN}}
\end{equation}

where $loss_{U{-}net}$ represents the loss function of U-net, $loss_{Mask \, R{-}CNN}$ denotes the loss function of Mask R-CNN, and $w_1$ along with $w_2$ are the respective weights for these loss functions.

To determine the weights for the MTL-02 loss function, we utilized the Dynamic Weighted Average (DWA) algorithm\cite{liu2019end}. The main idea is that tasks differ in their difficulty and speed of learning. Instead of giving each task a separate learning rate, it is better to adjust the weights dynamically so that all tasks advance at a similar pace. In simple terms, tasks that show a faster decrease in loss receive smaller weights, while those with a slower decrease are allocated larger weights.

\begin{equation}\label{19}
    {w}_k(t-1)=\frac{L_k(t-1)}{L_k(t-2)}
\end{equation}

\begin{equation}\label{20}
    {\lambda}_k(t)=\frac{K\cdot e^{\frac{{w}_k(t-1)}{T}}}{\sum_{i=1}^{K}e^{\frac{{w}_i(t-1)}{T}}}
\end{equation}
where $K$ is the number of tasks, $\lambda_k(t)$ is the weight of the task k at the i-th iteration, and $L_k(t)$ is calculated as the average loss in each epoch over several iterations. When $T$ is a temperature, $T=1$ indicates that $\lambda$ is equal to the result of softmax; However, as $T$ increases significantly, $\lambda$ approaches 1, leading to equal loss weights for all tasks.

%%%%%%%%%%%%%%Methods%%%%%%%%%%%%%%%%%

%%%%%%%%%%%%%%Experiments%%%%%%%%%%%%%%%%%
\section{Experiments}

\subsection{Research area and data}
\subsubsection{Research area}
\begin{figure}[h]
\centering %图片居中
\includegraphics[width=0.5\textwidth]{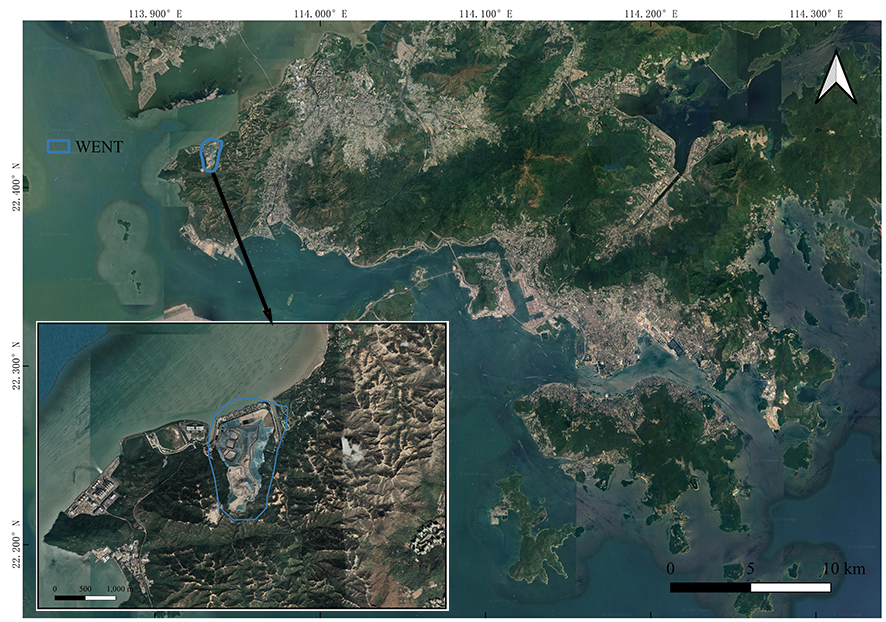} 
\caption{Research area}
  \small\centering
 The research site is the WENT landfill situated in the Tuen Mun District of Hong Kong SAR. The landfill is located on a hillside close to Shenzhen Bay, as indicated by the blue rectangle on the map.
\label{fig:picture002}
\end{figure}

Situated in the southern part of China, Hong Kong is predominantly marked by hilly terrain. The region, which is highly urbanized, experiences a humid and rainy climate, with an average annual precipitation exceeding 2000 millimeters. Within Hong Kong, there are two key landfills: West New Territories (WENT) landfill and North East New Territories (NENT) landfill. Since most of the PRISMA images were obtained from the WENT region, our generalization verification is mainly concentrated in this area. The Tuen Mun landfill (Fig. \ref{fig:picture002}), which spans approximately 27 hectares, is located in the southwest of the Tuen Mun district near Shenzhen Bay. It is one of the limited waste disposal sites sanctioned by the government in the Hong Kong Special Administrative Region and primarily receives municipal solid waste from various areas, including Hong Kong Island, Kowloon and New Territories.

\subsubsection{Hyperspectral Data}

The PRISMA satellite was launched on 22 March 2019, and offers hyperspectral images of global coverage with a spatial resolution of 30 m. The spectral smile is less than 5 $nm$, the spectral resolution is better than 12 $nm$  in a spectral range of 400-2500 nm (VNIR and SWIR regions) \cite{cogliati2021prisma}, and datasets are open-access. We have collected several PRISMA images of WENT, which are: $20220102$, $20220119$, $20220914$, $20230217$, $20230301$, $20230307$, $20230416$. These raw images will be used to verify the generalizability of deep learning models\cite{thorpe2013high}\cite{foote2020fast}.
%-------------------Figure2

%-------------------Figure2

 The EnMAP satellite was launched on 1 April 2022, and provides hyperspectral images with global coverage and a spatial resolution of 30 m. The spectral smile is less than 5 $nm$, and the spectral resolution is better than 12 $nm$ within a spectral range of 420-2450 nm (VNIR and SWIR regions). The EnMAP dataset originates from the publicly accessible atmospheric correction dataset as referenced in \cite{fuchs2023hyspecnet}, which comprises 11,000 images in total. These images are in Geotiff format, each with a resolution of $30\text{m}\times30\text{m}$ and containing 224 spectral bands in total. Initially, we extracted the data matrices along with the associated metadata from the GeoTIFF files. Subsequently, we eliminated the bands sensitive to water vapor and carbon dioxide, reducing the total to 41 bands. We then resampled these 41-band data to dimensions of $256\times256$ pixels using the nearest neighbor interpolation method. Throughout this process, no fitting was applied to the spectral curves. This procedure effectively enhances the area coverage of ground features, while preserving the original spatial resolution of $30 \text{m}$.

\subsection{Generation of diversified plumes in terms of emission rates and wind speeds}

\begin{figure*}[hb]%H为当前位置，!htb为忽略美学标准，htbp为浮动图形
\centering %图片居中
\includegraphics[width=1\textwidth]{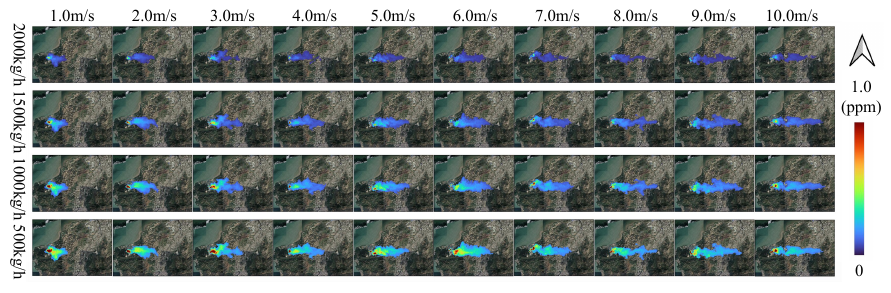} 
\caption{Snapshot of a methane plume simulated by the Palm software}
  \small\centering
This figure illustrates the methane plume at various emission rates and wind speeds, modeled with the Palm software. Each image was taken 180 minutes after the initiation of the emission, with simulation parameters specified in Section II.B.

\label{fig:picture003}
\end{figure*}

The simulated plumes were generated using Palm software\cite{2001PALM}\cite{maronga2015parallelized},using the average atmospheric pressure at sea level in Hong Kong (0.96 hPa) and a temperature of 296K. A specialized smog mode was implemented, including a methane dispersion feature, to simulate the diffusion of methane point sources within the troposphere from 0 to 10,000 meters. We defined four gradients for the methane point source emission rate: 500kg/h, 1000kg/h, 1500kg/h, and 2000kg/h. The wind direction was set from west to east, with wind speed gradients of 1m/s, 2m/s, 3m/s, 4m/s, 5m/s, 6m/s, 7m/s, 8m/s, 9m/s, and 10m/s (Fig. \ref{fig:picture003}). For each wind speed gradient, the methane point source emission rate gradients were simulated in sequence. The simulation period covered 0 to 2.5 hours, with snapshots taken at 30-second intervals from 1 to 2.5 hours, resulting in a total of 7200 plume snapshots. Among them, the simulated plume emission data with wind speeds of $2m/s$, and $10m/s$ were used to create the validation dataset, the simulated plume emission data with wind speeds of $1m/s$, and $9m/s$ were used to create the test dataset, while the remaining plume emission data were used to create the training dataset.

%-------------------Fig3
%-------------------Fig3

\subsection{Generation of multi-task learning datasets}

\begin{figure*}[hb]%H为当前位置，!htb为忽略美学标准，htbp为浮动图形
\centering %图片居中
\includegraphics[width=1\textwidth]{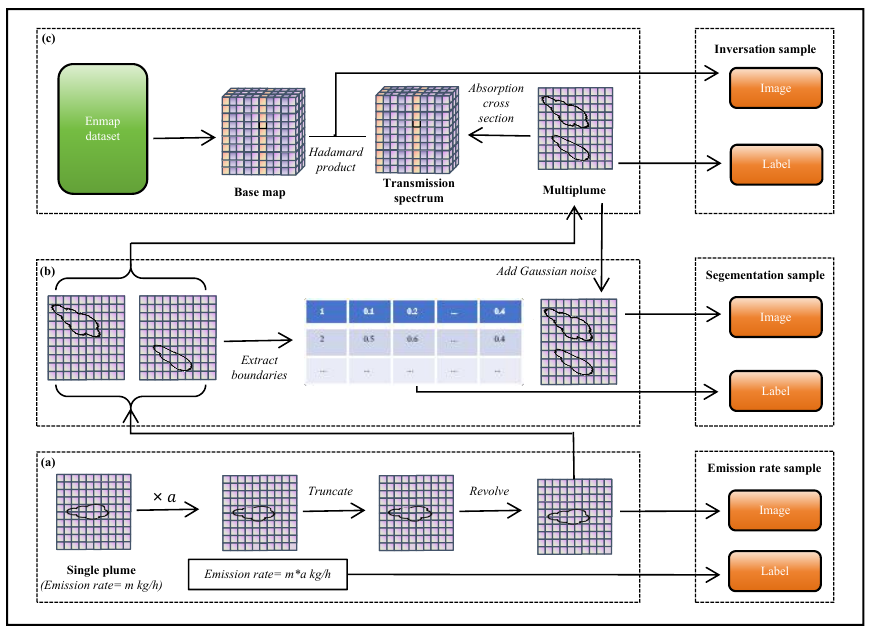} 
\caption{Generation process of training samples} 
  \small\centering
The training dataset is composed of three types of samples: Emission rate sample, Segmentation sample, and Inversion sample. The training dataset consists of numerous examples of these types of samples. (a) A snapshot of a plume, simulated by Palm, is scaled (including the emission rate), and concentration values below a certain threshold are set to zero. A random rotation from -170° to 170° is applied to produce a single enhanced plume, which acts as the emission rate sample image. The scaled emission rate serves as its label. (b) Up to N ($N\leq3$) plumes generated in step (a) are randomly merged (ensuring that overlap does not exceed 70\%) to create a multi-plume image. Noise from mag1c inversion outcomes is added to this multi-plume image to generate the segmentation sample image, with the boundary of each plume noted as the label. (c) Following the method in (a), a transmission spectrum matrix that matches the EnMAP base map dimensions is generated based on the multi-plume image from (b). The simulated EnMAP image is then created by applying the Hadamard product of the transmission spectrum matrix and the EnMAP base map, which serves as the inversion sample. The multi-plume image serves as the label for the inversion sample.
\label{picture0004}
\end{figure*}

The dataset we are going to create consists of three subsets: inversion dataset, segmentation dataset, and emission rate dataset. Each subset comprises three groups of data, namely Train, Val, and Test, which are respectively used for training, validation, and testing. Each group of data contains several samples, and each sample includes an image and its corresponding label. We assign plume snapshots of different wind speed gradients to Train, Val, and Test. The base map is also randomly divided into three groups and allocated to Train, Val, and Test accordingly. The creation process of a single sample group is as follows(Fig. \ref{picture0004}):

By capturing an image of a plume and averaging the methane concentrations at various heights, a 2D representation of a single plume with an emission rate of $m$ kg/h is produced. Scaling all pixel values in this 2D plume by a factor of $a$ results in a new emission rate of $m \times a$ kg/h, which is used as the sample label in the emission rate dataset. To ensure plume diversity, we randomly generate a threshold between 0.05 and 0.10, and set pixels in the 2D plume below this threshold to zero. A rotation enhancement ranging from -170° to 170° is then applied to the 2D plume, creating the image counterpart for the emission rate dataset sample. Plume segmentation is treated as an instance segmentation task. Initially, we create a blank $256 \times 256$ image filled with zeros. We then randomly select $N$ images from the emission rate dataset (where $N \leq 3$ and $N$ can be 0) and superimpose them onto the all-zero image to form a multi-plume image. Concurrently, we determine the boundaries of each plume and assign corresponding labels to the segmentation dataset. If $N=0$, the dataset will lack labels. Lastly, we randomly generate $256 \times 256$ Gaussian white noise and add it to the multi-plume image\cite{jongaramrungruang2022methanet}\cite{gil2002efficient} to create the image for the segmentation dataset. Additionally, the overlap ratio between any two plumes must not exceed 15%.

\begin{equation}\label{21}
 Overlapratio(A,B) = \frac{\sum_{x_i \in A \cap B} x_i}{\sum_{x_i \in A \cup B} x_i}
\end{equation}
where A,B represent two 2D plumes, $x_i$ is the pixel value in the plume.

The inversion data originate from multi-plume images created during the process of generating the segmentation dataset. These multi-plume images serve as labels for the inversion dataset samples. First, a multi-plume image is obtained and inspected for any emptiness. If found to be empty, a random EnMAP image is chosen as the sample image. On the other hand, if the multi-plume image is not empty, methane transmission spectra for each pixel value are constructed on a column-by-column basis. For columns that are entirely zeros, the transmission spectrum is defined as a $256 \times 1 \times 41$ matrix where all elements are set to 1. For columns with non-zero values, an optical depth matrix with dimensions $256 \times 1 \times K$ (where $K$ is the number of elements in the absorption cross-section dataset) is created using the approach explained in Section II.A. The negative logarithm of the pixel values in the optical depth matrix is then used to compute the original transmittance matrix.

First, the central wavelengths and half-bandwidths of the 41 EnMAP bands are determined to construct the spectral response function (SRF)\cite{green1998spectral}\cite{2008Toward}. Then, the SRF is utilized for normalized convolution along the $K$ dimension of the initial transmittance matrix, resulting in a $256 \times 1 \times 41$ 'column' transmission spectrum matrix. Following the processing of each column of the non-empty multi-plume image, all the transmission spectrum matrices are combined sequentially in their original column order to form a $256 \times 256 \times 41$ transmission spectrum matrix.

In the final step, the transmission spectrum matrix is merged with random EnMAP data through the Hadamard product to produce the image for the non-empty multi-plume. It's important to note that, since the simulated methane plumes are mainly below 3000 meters in altitude, we use the average temperature and atmospheric pressure of Hong Kong to determine the methane absorption cross-section and calculate the optical depth based on the mean atmospheric density from 0 to 3000 meters. While this method may reduce some accuracy in spectral simulation, the compromise is worthwhile given the substantial computational load of creating the inversion dataset. The entire code for generating the inversion dataset is implemented in PyTorch and executed on an NVIDIA GeForce RTX 3060 GPU.

%-------------------Fig2
%-------------------Fig2

\subsection{Accuracy Assessment}

\subsubsection{Regression Task}

Both inversion of methane concentration and estimation of emission rate are integral to the regression mission. Hence, to evaluate accuracy, we employed the RMSE and MAE metrics.

\begin{equation}\label{22}
    RMSE=\sqrt{\frac{1}{n} \sum_{i=1}^{n} (y_i - \hat{y_i})^2} 
\end{equation}
%RMSE

\begin{equation}\label{23}
    MAE=\frac{1}{n} \sum_{i=1}^{n} |y_i - \hat{y_i}| 
\end{equation}
%MAE
where $y$ is the ground truth value, $\hat{y}$ is the predicted value, and $n$ is the number of samples.

\subsubsection{Segmentation Task}
 % Methane plume segmentation falls under the purview of instance segmentation task. In an ideal scenario with complete supervision, metrics like IOU would be employed to gauge the accuracy of segmentation. However, due to the unavailability of precise labels, we rely on the methane instance detection rate and the performance of the segmented methane plume in RseNet as indicators of the segmentation task's effectiveness.
 
 %这一部分回头重写
Average precision (AP) is used to measure the accuracy of the methane plume segmentation task. The formula for calculating AP is as follows\cite{he2018hashing}:
 
1. Set a threshold and determine the number of $TP$, $FP$, $TN$, and $FN$ according to the threshold.

$TP$: True positive, the number of positive samples correctly classified as positive.

$FP$: False positive, the number of negative samples falsely classified as positive.

$TN$: True negative, the number of negative samples correctly classified as negative.

$FN$: False negative, the number of positive samples falsely classified as negative.

2. Calculate precision by dividing the number of true positives by the sum of the number of true positives and false positives.

\begin{equation}\label{24}
Precision = \frac{TP}{TP + FP}
\end{equation}

3. Calculate the recall by dividing the number of true positives by the sum of the number of true positives and false negatives.

\begin{equation}\label{25}
Recall = \frac{TP}{TP + FN} 
\end{equation}

4. Repeat this process for different thresholds to obtain a set of precision and recall values.

5. Calculate the AP value:
\begin{equation}\label{26}
AP=\sum_{i=1}^{n-1}(R_{i+1}-R_i)P_{inter}(R_i+1)
\end{equation}
where $(R_{i+1}-R_i)$ represents the span of the recall interval, and $P_{inter}(R_i+1)$ denotes the interpolated precision at a recall of $R_i+1$. The AP value is computed by aggregating all interpolated precision values, each weighted by their respective recall interval lengths between consecutive recalls.

%%%%%%%%%%%%%%Experiments%%%%%%%%%%%%%%%%%

%%%%%%%%%%%%%%Methods%%%%%%%%%%%%%%%%%%%%%%%%%%
\section{Results and Discussions}

\subsection{Simulation}
\begin{figure}[h]
\centering %图片居中
\includegraphics[width=0.5\textwidth]{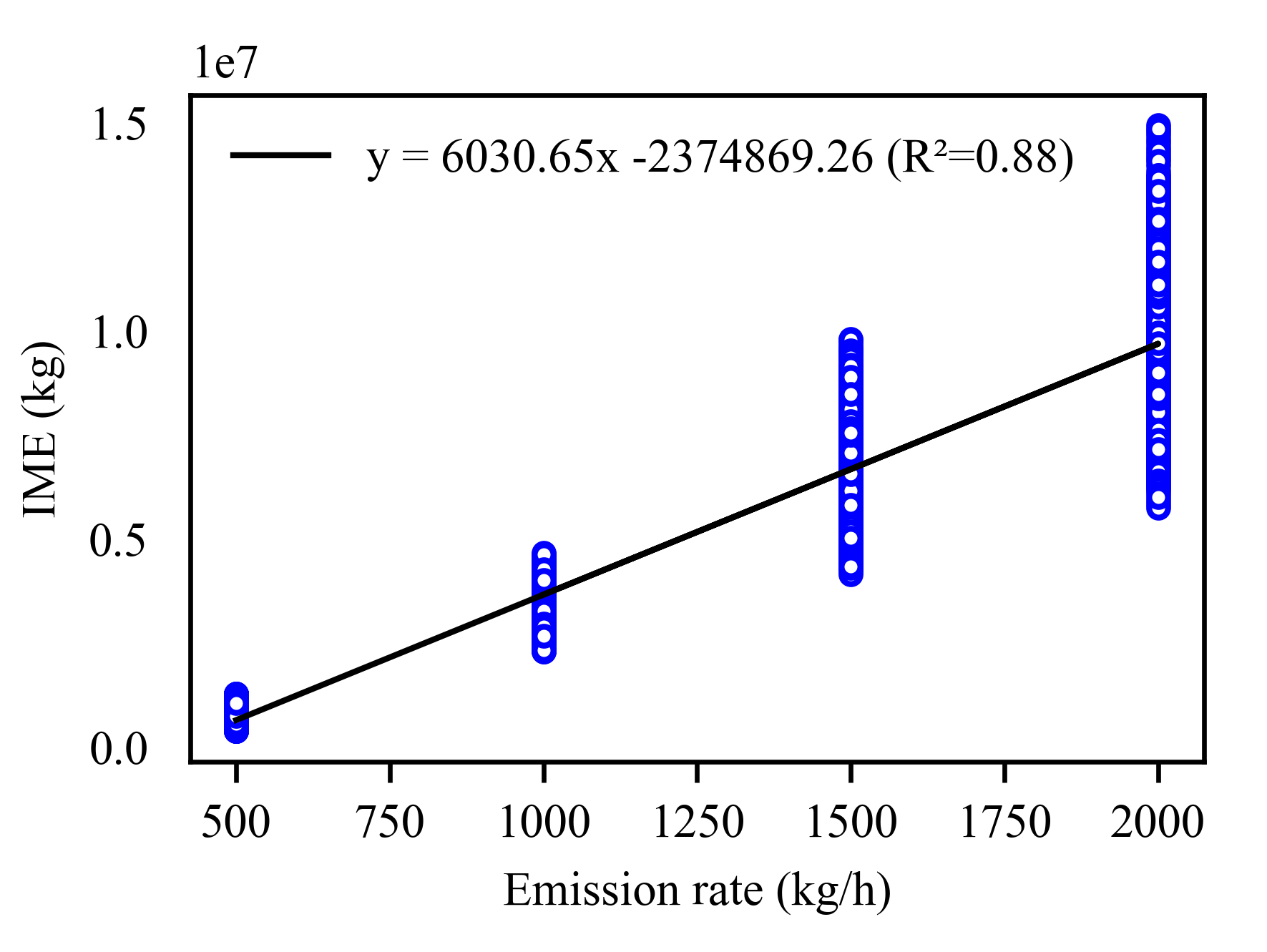} 
\caption{Correlation between emission rates and simulated plume dataset IME values.}
  \small\centering
  
We analyzed all 2D plume images generated by the PALM software and determined their IME values. A scatter plot was created to display the IME values of the plume images against their corresponding emission rates, and a linear fit was applied. The findings indicate a strong correlation between the IME values and emission rates of these plume images.  
\label{fig:picture005}
\end{figure}

To verify the accuracy of simulated plumes, we consulted relevant studies\cite{gorrono2023understanding}. The integrated methane emission (IME) values for a typical methane plume should exhibit a linear relationship with its emission rate. We computed the IME values for all 2D plume snapshots, plotted them against respective emission rates in a scatter plot, and fitted a linear curve, from which the coefficient of determination is derived. The findings show that the coefficient of determination for the linear curve was 0.88 (Fig. \ref{fig:picture005}), indicating a linear correlation between the IME values and emission rates of these 2D plume snapshots. Therefore, we conclude that the quality of simulated methane plumes is reasonable to a large extent. 
%-------------------Figure2
%-------------------Figure2

\begin{figure}[h]
\centering %图片居中
\includegraphics[width=0.5\textwidth]{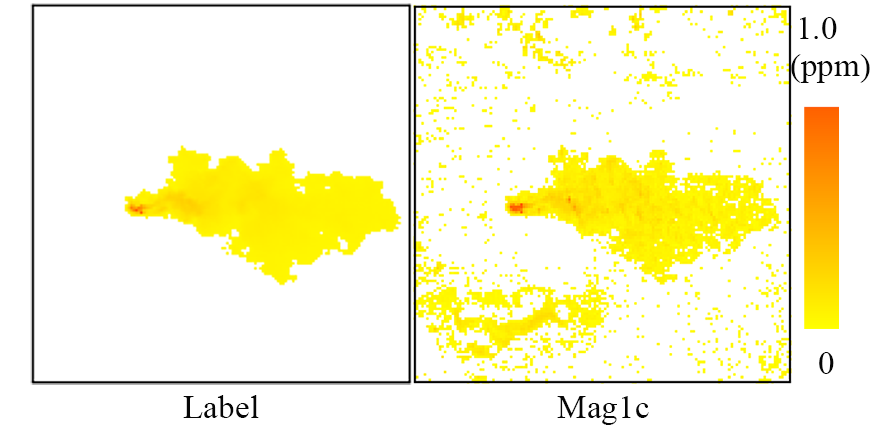} 
\caption{The results of end-to-end verification for simulation.}
  \small\centering

    Due to the absence of the ground truth value of methane concentration, the evaluation is performed through an end-to-end method: comparing the methane plume extracted from the simulated EnMAP image using mag1c with the methane plume added to the EnMAP base map. The results show that the methane plume extracted by mag1c closely resembles the plume incorporated into the EnMAP base map, indicating that the addition of the plume signal is appropriate.
  
\label{fig:picture006}
\end{figure}

We utilize simulation methods to create EnMAP datasets that include plume signals. Although the production process follows theoretical principles, the physical model is a simplified approximation of real-world conditions. Therefore, it is crucial to verify the accuracy of incorporating plume signals into the EnMAP base map. Without actual labels, we apply an end-to-end approach to validate the simulated EnMAP images. First, a random 2D plume snapshot is taken and a transmittance matrix is created following the method described in Section II. Next, a random EnMAP base map is chosen and combined with the transmittance matrix using the Hadamard product to generate a simulated EnMAP image. Following this, the Mag1c tool is employed to extract the methane concentration distribution map from the simulated EnMAP image. Finally, we compare the 2D plume snapshot with the methane concentration distribution map. If their methane concentration distributions match closely, the plume signal incorporation process can be deemed valid. In Fig. \ref{fig:picture006}, the left image shows a 2D plume snapshot, while the right image presents the methane concentration distribution map derived by Mag1c from the relevant EnMAP data. It is notable that the right image contains some additional noise compared to the left image, but the primary features are quite similar, thereby validating the plume signal incorporation process.

%-------------------Figure2
%-------------------Figure2

\subsection{Methane Concentration Inversion}

Each model was assessed on the test dataset for methane concentration inversion (see Table \ref{tab:my_table01}). The results from the validation suggest that the deep learning model consistently surpasses Mag1c in inversion accuracy, with U-net employing MSE yielding better precision than with SmoothL1. Reviewing the comprehensive validation results, it can be deduced that the U-net network exhibits superior predictive capabilities compared to Mag1c. However, the validation using simulated data may introduce some bias. To address this, we further evaluated U-net's generalization performance using PRISMA data from WENT.%_____________________________table
\noindent
\begin{table}[!htbp]
\centering
\caption{VALIDATION RESULTS OF METHANE CONCENTRATION INVERSION}
\label{tab:my_table01}
\begin{tabular}{|c|c|c|}
\hline
Methods& RMSE$\slash ppm$ & MAE$\slash ppm$ \\
\hline
Mag1c &	 0.5377& 0.1392\\
\hline
U-net + Smooth $L_1$ loss&	0.1208&  0.0225\\
\hline
U-net + MSE loss& 0.0926&	\textbf{0.0159}\\
\hline
MTL-02&	 \textbf{0.0790}&	0.0196\\
\hline
\end{tabular}
\end{table}
%_____________________________table

\begin{figure}[h]
\centering %图片居中
\includegraphics[width=0.5\textwidth]{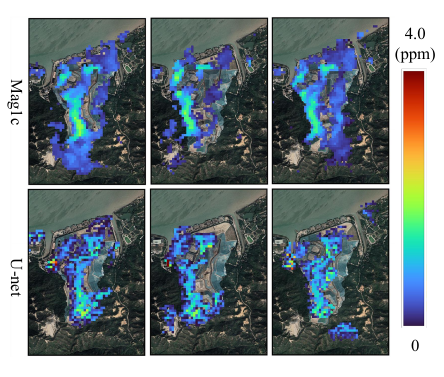} 
\caption{Comparison of generalization results between
U-net and Mag1c on real data.}
  \small\centering
    Sample illustrations of U-net's generalization performance on PRISMA image.

\label{fig:picture007}
\end{figure}

After eliminating the sensitive bands for water vapor and carbon dioxide, PRISMA and EnMAP retained 49 and 41 bands, respectively. The PRISMA data were then resampled to 41 bands in the spectral dimension, aligned with the center wavelengths of the 41 EnMAP bands used in the prediction. Subsequently, the pre-trained U-net was applied to invert the methane plume in the area, and its results were compared to the Mag1c inversion based on the 49 PRISMA bands. As depicted in Fig. \ref{fig:picture007}, three representative inversion visualizations were presented. It is observable that U-net and Mag1c provide consistent inversion results in certain datasets, while some differences are evident in others. Due to the lack of true methane plume concentration values in the WENT region, determining the most accurate method is challenging. We investigated potential reasons for inconsistencies between the two approaches: 1. The uneven terrain of the WENT region and proximity to the ocean create a complex noise distribution in the PRISMA data, affecting the methods differently. 2. The PRISMA data resampling process results in some plume information loss, leading to disparities in U-net and Mag1c inversion results. 3. The limited number of EnMAP base maps used for U-net training in rugged areas may cause U-net to predict methane column concentrations based on smoother terrain.
%-------------------Figure2
%-------------------Figure2

\subsection{Methane Plume Segmentation}

The Mask R-CNN model was trained on two versions of the plume segmentation dataset as detailed in Section II.C. One version used simulated hyperspectral images (hyperspectral image) as input, while the other used methane plume distribution maps (single). Despite this, the labels for both versions of the data set were identical. Similarly, model validation was performed separately using validation sets corresponding to the two versions of the data set, maintaining identical labels. Note that the Active Contour method was validated exclusively using single methane plume maps as input because it is an unsupervised image segmentation technique that requires an explicit relationship between pixel values and task-dependent physical properties. The results indicated that the Mask R-CNN model achieved higher precision than the Active Contour method (Table \ref{tab:my_tabel02}). However, when assessing the AP metric at $AP_{95}$, the Active Contour method outperformed Mask R-CNN, suggesting that Active Contour excels in isolating individual plume masks. Despite this, the overall performance of Active Contour is inferior to that of Mask R-CNN due to its difficulty in effectively distinguishing overlapping plumes and its tendency to create small, extraneous segmented areas.

%_____________________________two-colume table
\noindent
\begin{table}[!htbp]
\centering
\caption{VALIDATION RESULTS OF METHANE PLUME SEGEMENTATION}
\label{tab:my_tabel02}
\begin{tabular}{|c|c|c|c|c|c|c|c|c|c|c|c|}
\hline
Models & $AP_{50}$ &$AP_{75}$	&$AP_{95}$	&$AP_{50:95}$ \\
\hline
Active Contour	&42.03	&39.04	&\textbf{9.35}	&34.96\\
\hline
MTL-01	&\textbf{84.06} &\textbf{79.93}&	0.00& \textbf{64.74}\\
\hline
Mask R-CNN single &	82.58 &	76.73 &	0.00&	61.84\\
\hline
Mask R-CNN hyper &29.28		&14.70	&0.00	&15.07\\
\hline
MTL-02& 35.21 &15.44	&	0.00&17.14		\\
\hline

\end{tabular}
\end{table}

\begin{figure*}[hb]%H为当前位置，!htb为忽略美学标准，htbp为浮动图形
\centering %图片居中
\includegraphics[width=1\textwidth]{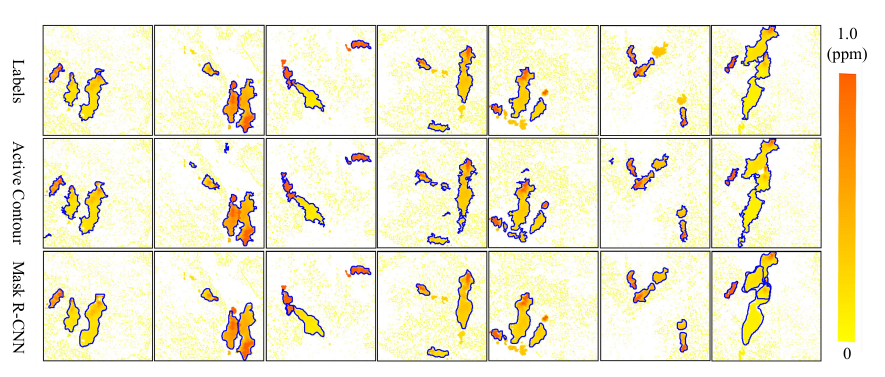} 
\caption{Segmentation results of various methods on the test dataset}
  \small\centering
The region outlined by the blue border shows the segmentation of methane plumes produced by different methods.
\label{fig:picture008}
\end{figure*}

Both versions of Mask R-CNN demonstrate superior precision in extracting the plume mask from the methane plume distribution map compared to extracting it directly from hyperspectral images. This is primarily due to two factors: first, the EnMAP data, produced through a two-step simulation process, inherently contain more errors than the one-step simulation method. Second, extraction of the EnMAP plume mask involves both inversion and segmentation steps, which increases the likelihood of inaccuracies compared to direct extraction from the methane plume distribution map. Regarding the "single" Mask R-CNN, despite its capability to efficiently segment overlapping plumes, its segmentation outcome still exhibits some errors (Fig. \ref{fig:picture008}). These errors can generally be classified as follows: 1. If a plume is divided into two sections, the single Mask R-CNN may incorrectly segment it into two separate plumes. 2. For overlapping plumes, the single Mask R-CNN may generate redundant or duplicate segmentations.
%-------------------Fig2
%-------------------Fig2

\begin{figure}[h]
\centering %图片居中
\includegraphics[width=0.5\textwidth]{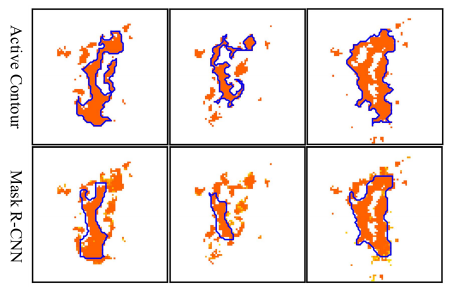} 
\caption{Comparison of generalization results of different methods on real plume images}
  \small\centering
To assess the generalizability of the Mask R-CNN model on real-world data, we employed mag1c to invert the methane concentration map derived from multiple PRISMA datasets in the WENT region. Subsequently, we used both Active Contour and Mask R-CNN to extract the methane plume. The area marked by the blue line in the figure indicates the extracted methane plume. It is evident that Active Contour aims to extract all connected pixel regions extensively, while Mask R-CNN focuses on pixel regions with a funnel-shaped distribution. Nonetheless, the overall count of methane plumes identified by both methods is comparable.
\label{fig:picture009}
\end{figure}

To evaluate the Mask R-CNN model's ability to generalize to real-world contexts, mag1c is used to reverse-engineer methane concentration maps from various PRISMA images within the WENT area. Subsequently, both Active Contour and Mask R-CNN were applied to delineate methane plumes. The blue boxes in Fig. \ref{fig:picture009} denote these identified plumes. It is evident that Active Contour maps all connected pixel regions, whereas Mask R-CNN typically highlights funnel-shaped pixel regions, indicating some resistance to noise. Despite this, both methods identified a comparable number of methane plumes. We acknowledge the absence of precise data on the exact count of methane emission sources (or plumes) in the WNET area. Nevertheless, visual inspection reveals that both techniques detected the major plumes and overlooked several smaller ones on the right. This omission could be attributed to the rugged coastal terrain of the WENT region, where plume dispersion is influenced by intricate factors like topography and sea-land breeze dynamics, resulting in more complex plume formations.

%-------------------Figure2
%-------------------Figure2

\subsection{Estimation of Methane Emission Rate}

We employed ResNet-50 and AlexNet to estimate emission rates by training them on the provided dataset and evaluating the accuracy of their estimations on the corresponding test set (Table \ref{tab:my_tabel03}). The results reveal that ResNet-50 outperforms the IME algorithm in the validation set. Although AlexNet’s estimation accuracy is slightly higher than that of the IME algorithm, it still does not match ResNet-50's performance. This aligns with the common understanding in deep learning that deeper networks enable each layer to address simpler tasks. As a result, more complex nonlinear representations are possible, which helps to learn intricate mappings\cite{2016On}, and boosts the generalization ability and robustness of the network.%\cite{Mont2014On}\cite{2007Scaling}
%_____________________________table
\noindent
\begin{table}[!htbp]
\centering
\caption{VALIDATION RESULTS OF EMISSION RATE ESTIMATION OF SINGLE NETWORKS}
\label{tab:my_tabel03}
\begin{tabular}{|c|c|c|}
\hline
Methods& RMSE$\slash kg*h^{-1}$ & MAE$\slash kg*h^{-1}$ \\
\hline
IME &	 187.47 & 142.05\\

\hline
AlexNet + MSE loss&	 177.31&141.03\\

\hline
 ResNet + MSE loss&\textbf{126.93}&\textbf{96.09}\\
\hline

\end{tabular}
\end{table}
%_____________________________table

\begin{figure}[h]
\centering %图片居中
\includegraphics[width=0.5\textwidth]{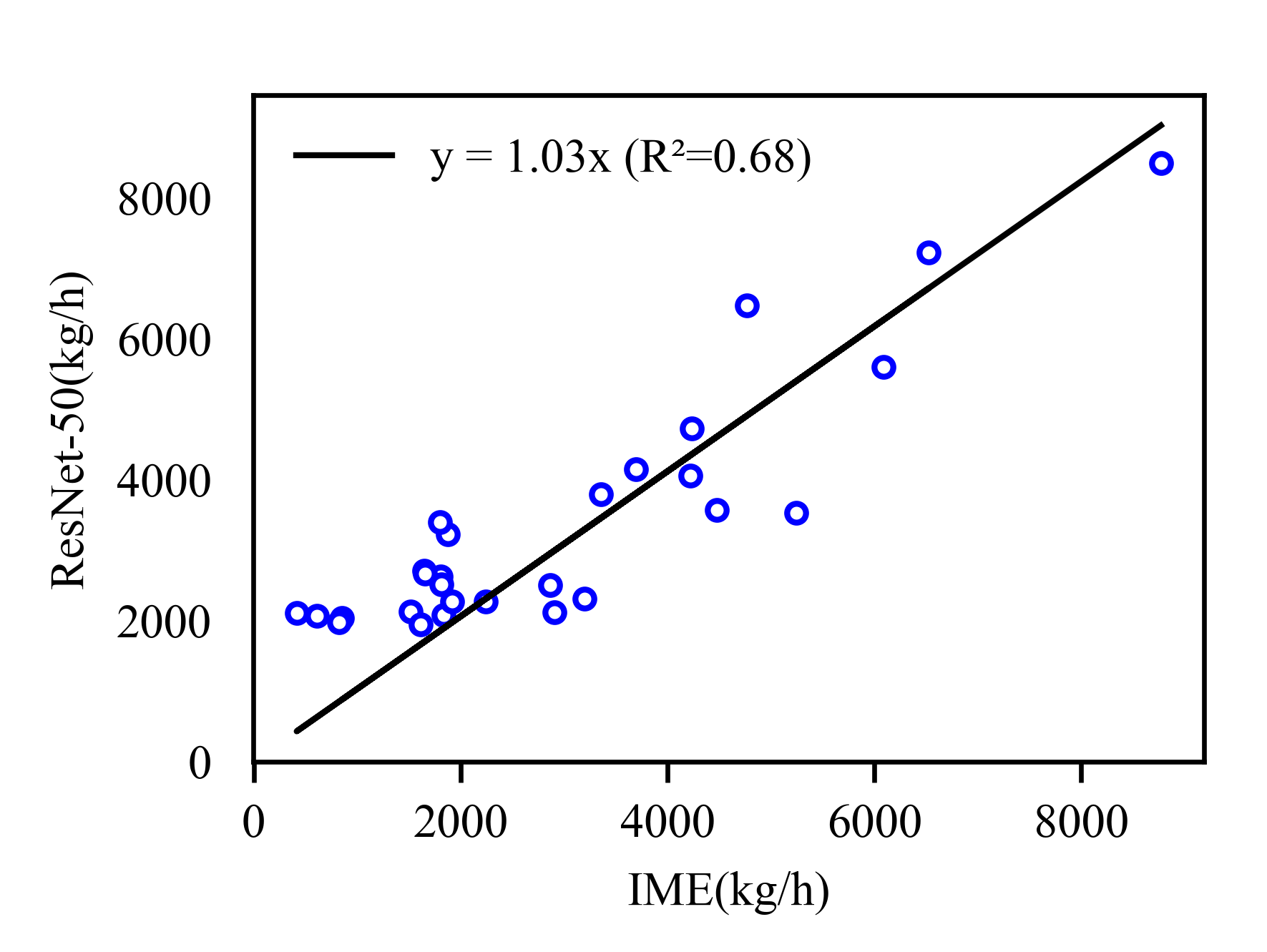} 
\caption{Comparison of generalization results between ResNet-50 and IME on real data}
  \small\centering
We manually extracted 28 methane plumes from the inversion results of PRISMA's WENT region and predicted their emission rates using the IME method and the trained ResNet-50 model separately. The prediction results are shown in the figure. Both methods yielded comparable results. However, it can also be observed that ResNet-50 shows some saturation in regions with low emission rates.

\label{fig:picture010}
\end{figure}

To assess ResNet-50's ability to generalize to real-world data, we employed Mag1c to invert methane concentration distribution maps derived from RPISMA data in the WENT region and manually identified 28 methane plumes. We estimated their emission rates using both the IME method and the pre-trained ResNet-50 model, constructed scatter plots, and performed linear curve fitting. As shown in Fig. \ref{fig:picture010}, the scatter plot illustrates that the majority points are aligned closely to a straight line, indicating that the predictions of both approaches are typically in good agreement with the real-world dataset. However, it is also apparent that the ResNet-50 model shows a tendency to saturate in areas with lower emission rates.
%-------------------Figure2
%-------------------Figure2

\subsection{Serial Network}

Although individual models demonstrated high accuracy for each specific task, the efficacy of serializing multi-network models still warrants further examination. Initially, we combined Mask R-CNN and ResNet-50 to handle the tasks of segmenting methane plumes from concentration maps and estimating their emission rates. We assessed the accuracy of these emission rate estimates in the validation set under three scenarios: 1. For true positive segmented plumes, we used pre-trained ResNet-50 to estimate emission rates and compared them with the label's emission rates. 2. For false-positive segmented plumes, we similarly used ResNet-50 to estimate their emission rates and compared them against 0 kg/h. 3. For misdetected plumes, we assumed their emission rate estimates to be 0 kg/h and compared these estimates with the corresponding labels. Ultimately, RMSE and MAE between all predicted emission rates and label values were calculated (Table \ref{tab:my_tabel04}). The results indicated that the accuracy of the emission rate estimation using Mask R-CNN + ResNet-50 was significantly lower than using ResNet-50 for single-task emission rate estimation. It indicates that the serialization of Mask R-CNN and ResNet-50 could introduce additional errors in the overall performance of methane flux rate estimation.
%_____________________________table
\noindent
\begin{table}[!htbp]
\centering
\caption{VALIDATION RESULTS OF EMISSION RATE ESTIMATION OF SERIAL NETWORKS}
\label{tab:my_tabel04}
\begin{tabular}{|l|c|c|}
\hline
Methods& RMSE$\slash kg*h^{-1}$ & MAE$\slash kg*h^{-1}$ \\
\hline
 ResNet-50 + MSE loss&\textbf{126.93}&\textbf{96.09}\\
\hline
 Mask R-CNN+ResNet $plume_{300}$ 	&1157.45	&859.56\\
\hline
MTL-01 	&1038.30	&788.31 \\
\hline

\end{tabular}

\end{table}
% _____________________________table

To investigate additional sources of error, we chose segmented plumes with significant errors (e.g., exceeding 500 kg/h). The analysis identified two additional error sources: 1. Errors stemming from small patches that are essentially noise; 2. Errors arising from false positive patches.

We examined the distribution of minor noise patches and discovered that the majority are defined by a mask size of less than 300 pixels, while authentic plume mask sizes seldom fall below 300 pixels. As a result, we exclude segmented masks smaller than 300 pixels in size and reassess the accuracy of Mask R-CNN + ResNet-50 for estimating emission rates.

\subsection{Multi-task Learning I (MTL-01)}

We trained the $MTL-01$ network, and after 30 epochs, $MTL-01$ outperformed Mask R-CNN in the plume segmentation task. As a result, the loss in estimating the emission rates of multiple methane sources was also lower than when using a serialized approach with $Mask \,R-CNN$ and $ResNet-50$. Additionally, by simulating a smaller training dataset with both methods, $MTL-01$ demonstrated even higher accuracy compared to the serialized training of $Mask\, R-CNN$ and $MTL-01$. It is evident that the separation of extra plumes is suppressed (Fig. \ref{fig:picture009}), which lowers the predicted emission rate. $MTL-01$ mainly segments the core regions of the jet stream while overlooking some smaller sections of the tail. This has led to improvements in both the plume segmentation task and the emission rate estimation (Table \ref{tab:my_tabel02}, Table \ref{tab:my_tabel04}).

%-------------------Figure2
\begin{figure}[h]
\centering %图片居中
\includegraphics[width=0.5\textwidth]{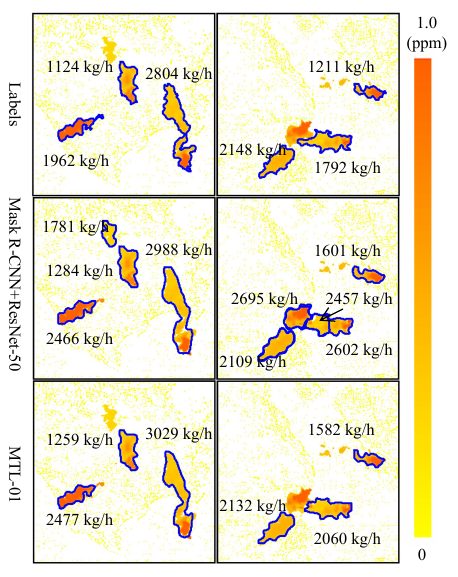} 
\caption{Comparison of the sequential network and multi-task learning network I (MTL-01).}
\label{fig:picture011}
\end{figure}
%-------------------Figure2

\subsection{Multi-task Learning II (MTL-02)}

We trained the MTL-02 model, which reached convergence after 30 epochs, showing a rapid decline in both loss function values and validation accuracy values. In the end, MTL-02 obtained slightly better accuracy in methane concentration inversion than U-net (Table \ref{tab:my_table01}), and superior accuracy in plume segmentation compared to Hyper Mask R-CNN (Table \ref{tab:my_tabel02}). This supports the notion that multi-task learning could enhance the convergence rate and fitting accuracy of the model, likely due to the shared information among various labels.

\subsection{Limitations}

Monitoring methane plumes is an intricate issue, and the current research in this domain is neither thoroughly explored nor systematic. Our proposed method also has some constraints: 1. The shapes of the simulated plumes that we used mostly adhere to a standard Gaussian model, which represents an ideal scenario. However, in reality, methane emissions from point sources are influenced by a variety of factors like topography and humidity, leading to non-Gaussian plumes. Whether our method applies to non-Gaussian plumes remains an open question; 2. The maximum overlap ratio of the plumes that we examined is $15\%$. However, naturally, many sources of methane emission are situated close to each other, and the overlap ratio of two plumes can be significantly higher than $15\%$. In such scenarios, even though we can generate highly overlapped methane plumes, it is challenging to provide precise training labels, necessitating the exploration of point-based semi-supervised learning techniques to achieve reasonable segmentation outcomes for these highly overlapped plumes; 3. The acquisition of hyperspectral satellite imagery is significantly affected by weather conditions, and hyperspectral image data sources are scarce in low-latitude coastal regions. Hence, a solution relying solely on hyperspectral images may not fulfill high-frequency monitoring needs. It is essential to devise an ensemble estimation method based on multiple data sources.

\section{Conclusion}

Currently, there are significant uncertainties in monitoring fugitive methane emissions using satellite imaging spectrometers. We have created a demonstrative method utilizing deep learning models and simulated hyperspectral images for methane concentration inversion, plume segmentation, and emission rate estimation in landfills. Specifically, U-net was employed for methane concentration inversion, yielding higher accuracy and reduced noise compared to mag1c. Mask R-CNN was used for plume segmentation, eliminating subjectivity in the process. For estimating methane plume emission rates, ResNet-50 is utilized to bypass the dependence on wind speed measurements. We also addressed additional errors in emission rate estimation caused by serial networks by developing a multi-task model (MTL-01) to reduce these errors and enhance the precision of plume segmentation. Furthermore, a multi-task model (MTL-02) that performs both methane concentration inversion and plume segmentation was proposed, achieving higher validation accuracy than individual single-task models. Finally, we highlighted the limitations of the current approach, emphasizing the need for future research on highly overlapped plumes, non-Gaussian plumes, and the integration of multi-source data for inversion.

% or


\appendices

\section*{Acknowledgment}

The work described in this paper was substantially supported by a grant from the ECF and EPD of the Hong Kong Special Administrative Region, China (Project No. PolyU ECF111/2020).

\ifCLASSOPTIONcaptionsoff
  \newpage
\fi



%\bibliographystyle{IEEEtran}

%%%%%%%%%%%%%%%%%%%%References%%%%%%%%%%%%%%%%%%%%%%%%%%
\printbibliography 

@article{magazzino2020relationship,
  title={The relationship between municipal solid waste and greenhouse gas emissions: Evidence from Switzerland},
  author={Magazzino, Cosimo and Mele, Marco and Schneider, Nicolas},
  journal={Waste Management},
  volume={113},
  pages={508--520},
  year={2020},
  publisher={Elsevier}
}

@article{zhang2019greenhouse,
  title={Greenhouse gas emissions from landfills: A review and bibliometric analysis},
  author={Zhang, Chengliang and Xu, Tong and Feng, Hualiang and Chen, Shaohua},
  journal={Sustainability},
  volume={11},
  number={8},
  pages={2282},
  year={2019},
  publisher={MDPI}
}

@article{bogner2007waste,
  title={Waste management in climate change 2007: Mitigation. contribution of working group iii to the fourth assessment report of the intergovernmental panel on climate change},
  author={Bogner, J and Ahmed, M Abdelrafie and Diaz, C and Faaij, A and Gao, Q and Hashimoto, S and Mareckova, K and Pipatti, R and Zhang, T},
  journal={Cambridge University Press, Cambridge},
  year={2007}
}

@misc{canadell2021global,
  title={Global carbon and other biogeochemical cycles and feedbacks},
  author={Canadell, Josep G and Monteiro, Pedro MS and Costa, Marcos H and Da Cunha, Leticia Cotrim and Cox, Peter M and Eliseev, Alexey V and Henson, Stephanie and Ishii, Masao and Jaccard, Samuel and Koven, Charles and others},
  year={2021}
}

@incollection{stocker2013technical,
  title={Technical summary},
  author={Stocker, Thomas F and Qin, Dahe and Plattner, G-K and Alexander, Lisa V and Allen, Simon K and Bindoff, Nathaniel L and Br{\'e}on, F-M and Church, John A and Cubasch, Ulrich and Emori, Seita and others},
  booktitle={Climate change 2013: the physical science basis. Contribution of Working Group I to the Fifth Assessment Report of the Intergovernmental Panel on Climate Change},
  pages={33--115},
  year={2013},
  publisher={Cambridge University Press}
}

@inproceedings{iravanian2020types,
  title={Types of contamination in landfills and effects on the environment: a review study},
  author={Iravanian, A and Ravari, Sh O},
  booktitle={IOP Conference Series: Earth and Environmental Science},
  volume={614},
  number={1},
  pages={012083},
  year={2020},
  organization={IOP Publishing}
}

@article{yang2021greenhouse,
  title={Do greenhouse gas emissions drive extreme weather conditions at the city level in China? Evidence from spatial effects analysis},
  author={Yang, Zhiming and Kagawa, Shigemi and Li, Jing},
  journal={Urban Climate},
  volume={37},
  pages={100812},
  year={2021},
  publisher={Elsevier}
}

@article{10.1115/1.3124648,
    author = {Alfonsi, Giancarlo},
    title = "{Reynolds-Averaged Navier–Stokes Equations for Turbulence Modeling}",
    journal = {Applied Mechanics Reviews},
    volume = {62},
    number = {4},
    pages = {040802},
    year = {2009},
    month = {06},
}

@article{chagovets2007effective,
  title={Effective kinematic viscosity of turbulent He II},
  author={Chagovets, TV and Gordeev, AV and Skrbek, L},
  journal={Physical Review E},
  volume={76},
  number={2},
  pages={027301},
  year={2007},
  publisher={APS}
}

@article{launder1975progress,
  title={Progress in the development of a Reynolds-stress turbulence closure},
  author={Launder, Brian Edward and Reece, G Jr and Rodi, W},
  journal={Journal of fluid mechanics},
  volume={68},
  number={3},
  pages={537--566},
  year={1975},
  publisher={Cambridge University Press}
}

@inproceedings{fuchs2023hyspecnet,
  title={Hyspecnet-11k: A large-scale hyperspectral dataset for benchmarking learning-based hyperspectral image compression methods},
  author={Fuchs, Martin Hermann Paul and Demir, Beg{\"u}m},
  booktitle={IGARSS 2023-2023 IEEE International Geoscience and Remote Sensing Symposium},
  pages={1779--1782},
  year={2023},
  organization={IEEE}
}

@article{sagaut2002large,
  title={Large eddy simulation for incompressible flows: An introduction. scientific computation series},
  author={Sagaut, Pierre and Lee, Yu-Tai},
  journal={Appl. Mech. Rev.},
  volume={55},
  number={6},
  pages={B115--B116},
  year={2002}
}

@article{pope2001turbulent,
  title={Turbulent flows},
  author={Pope, Stephen B},
  journal={Measurement Science and Technology},
  volume={12},
  number={11},
  pages={2020--2021},
  year={2001}
}

@article{germano1991dynamic,
  title={A dynamic subgrid-scale eddy viscosity model},
  author={Germano, Massimo and Piomelli, Ugo and Moin, Parviz and Cabot, William H},
  journal={Physics of Fluids A: Fluid Dynamics},
  volume={3},
  number={7},
  pages={1760--1765},
  year={1991},
  publisher={American Institute of Physics}
}

@article{domaradzki1997subgrid,
  title={A subgrid-scale model based on the estimation of unresolved scales of turbulence},
  author={Domaradzki, J Andrzej and Saiki, Eileen M},
  journal={Physics of Fluids},
  volume={9},
  number={7},
  pages={2148--2164},
  year={1997},
  publisher={American Institute of Physics}
}

@article{cogliati2021prisma,
  title={The PRISMA imaging spectroscopy mission: Overview and first performance analysis},
  author={Cogliati, Sergio and Sarti, F and Chiarantini, L and Cosi, M and Lorusso, R and Lopinto, E and Miglietta, F and Genesio, L and Guanter, Luis and Damm, Alexander and others},
  journal={Remote Sensing of Environment},
  volume={262},
  pages={112499},
  year={2021},
  publisher={Elsevier}
}

@article{2001PALM,
  title={PALM-A large-eddy simulation model performing on massively parallel computers},
  author={ Raasch, Siegfried  and  Sch, Michael },
  journal={Meteorologische Zeitschrift},
  volume={10},
  number={5},
  pages={363-372},
  year={2001},
}

@article{maronga2015parallelized,
  title={The Parallelized Large-Eddy Simulation Model (PALM) version 4.0 for atmospheric and oceanic flows: model formulation, recent developments, and future perspectives},
  author={Maronga, Bj{\"o}rn and Gryschka, Micha and Heinze, Rieke and Hoffmann, Fabian and Kanani-S{\"u}hring, Farah and Keck, Marius and Ketelsen, K and Letzel, Marcus Oliver and S{\"u}hring, Matthias and Raasch, Siegfried},
  journal={Geoscientific Model Development},
  volume={8},
  number={8},
  pages={2515--2551},
  year={2015},
  publisher={Copernicus GmbH}
}

@article{gorrono2023understanding,
  title={Understanding the potential of Sentinel-2 for monitoring methane point emissions},
  author={Gorro{\~n}o, Javier and Varon, Daniel J and Irakulis-Loitxate, Itziar and Guanter, Luis},
  journal={Atmospheric Measurement Techniques},
  volume={16},
  number={1},
  pages={89--107},
  year={2023},
  publisher={Copernicus GmbH}
}

@article{guanter2021mapping,
  title={Mapping methane point emissions with the PRISMA spaceborne imaging spectrometer},
  author={Guanter, Luis and Irakulis-Loitxate, Itziar and Gorro{\~n}o, Javier and S{\'a}nchez-Garc{\'\i}a, Elena and Cusworth, Daniel H and Varon, Daniel J and Cogliati, Sergio and Colombo, Roberto},
  journal={Remote Sensing of Environment},
  volume={265},
  pages={112671},
  year={2021},
  publisher={Elsevier}
}

@article{green1998spectral,
  title={Spectral calibration requirement for Earth-looking imaging spectrometers in the solar-reflected spectrum},
  author={Green, Robert O},
  journal={Applied optics},
  volume={37},
  number={4},
  pages={683--690},
  year={1998},
  publisher={Optica Publishing Group}
}

@article{2008Toward,
  title={Toward scene-based retrieval of spectral response functions for hyperspectral imagers using Fraunhofer features},
  author={ Brazile, Jason  and  Neville, Robert A  and  Staenz, Karl  and  Schl?Pfer, Daniel  and  Sun, Lixin  and  Itten, Klaus I },
  journal={Canadian Journal of Remote Sensing},
  volume={34},
  number={sup1},
  pages={S43-S58},
  year={2008},
}

@article{foote2020fast,
  title={Fast and accurate retrieval of methane concentration from imaging spectrometer data using sparsity prior},
  author={Foote, Markus D and Dennison, Philip E and Thorpe, Andrew K and Thompson, David R and Jongaramrungruang, Siraput and Frankenberg, Christian and Joshi, Sarang C},
  journal={IEEE Transactions on Geoscience and Remote Sensing},
  volume={58},
  number={9},
  pages={6480--6492},
  year={2020},
  publisher={IEEE}
}

@article{jongaramrungruang2022methanet,
  title={MethaNet--An AI-driven approach to quantifying methane point-source emission from high-resolution 2-D plume imagery},
  author={Jongaramrungruang, Siraput and Thorpe, Andrew K and Matheou, Georgios and Frankenberg, Christian},
  journal={Remote Sensing of Environment},
  volume={269},
  pages={112809},
  year={2022},
  publisher={Elsevier}
}

@article{gil2002efficient,
  title={Efficient dilation, erosion, opening, and closing algorithms},
  author={Gil, Joseph Yossi and Kimmel, Ron},
  journal={IEEE Transactions on Pattern Analysis and Machine Intelligence},
  volume={24},
  number={12},
  pages={1606--1617},
  year={2002},
  publisher={IEEE}
}

@article{thorpe2013high,
  title={High resolution mapping of methane emissions from marine and terrestrial sources using a Cluster-Tuned Matched Filter technique and imaging spectrometry},
  author={Thorpe, Andrew K and Roberts, Dar A and Bradley, Eliza S and Funk, Christopher C and Dennison, Philip E and Leifer, Ira},
  journal={Remote Sensing of Environment},
  volume={134},
  pages={305--318},
  year={2013},
  publisher={Elsevier}
}

@article{foote2021impact,
  title={Impact of scene-specific enhancement spectra on matched filter greenhouse gas retrievals from imaging spectroscopy},
  author={Foote, Markus D and Dennison, Philip E and Sullivan, Patrick R and O'Neill, Kelly B and Thorpe, Andrew K and Thompson, David R and Cusworth, Daniel H and Duren, Riley and Joshi, Sarang C},
  journal={Remote Sensing of Environment},
  volume={264},
  pages={112574},
  year={2021},
  publisher={Elsevier}
}

@article{wagner2004max,
  title={MAX-DOAS O4 measurements: A new technique to derive information on atmospheric aerosols—Principles and information content},
  author={Wagner, T and Dix, B v and Friedeburg, C v and Frie{\ss}, U and Sanghavi, S and Sinreich, R and Platt, U},
  journal={Journal of Geophysical Research: Atmospheres},
  volume={109},
  number={D22},
  year={2004},
  publisher={Wiley Online Library}
}

@book{platt2008differential,
  title={Differential absorption spectroscopy},
  author={Platt, Ulrich and Stutz, Jochen and Platt, Ulrich and Stutz, Jochen},
  year={2008},
  publisher={Springer}
}

@article{thompson2015atmospheric,
  title={Atmospheric correction for global mapping spectroscopy: ATREM advances for the HyspIRI preparatory campaign},
  author={Thompson, David R and Gao, Bo-Cai and Green, Robert O and Roberts, Dar A and Dennison, Philip E and Lundeen, Sarah R},
  journal={Remote Sensing of Environment},
  volume={167},
  pages={64--77},
  year={2015},
  publisher={Elsevier}
}

@article{thorpe2017airborne,
  title={Airborne DOAS retrievals of methane, carbon dioxide, and water vapor concentrations at high spatial resolution: application to AVIRIS-NG},
  author={Thorpe, Andrew K and Frankenberg, Christian and Thompson, David R and Duren, Riley M and Aubrey, Andrew D and Bue, Brian D and Green, Robert O and Gerilowski, Konstantin and Krings, Thomas and Borchardt, Jakob and others},
  journal={Atmospheric Measurement Techniques},
  volume={10},
  number={10},
  pages={3833--3850},
  year={2017},
  publisher={Copernicus GmbH}
}

@article{theiler2006effect01,
  title={Effect of signal contamination in matched-filter detection of the signal on a cluttered background},
  author={Theiler, James and Foy, Bernard R},
  journal={IEEE Geoscience and Remote Sensing Letters},
  volume={3},
  number={1},
  pages={98--102},
  year={2006},
  publisher={Ieee}
}

@article{funk2001clustering,
  title={Clustering to improve matched filter detection of weak gas plumes in hyperspectral thermal imagery},
  author={Funk, Christopher C and Theiler, James and Roberts, Dar A and Borel, Christoph C},
  journal={IEEE transactions on geoscience and remote sensing},
  volume={39},
  number={7},
  pages={1410--1420},
  year={2001},
  publisher={IEEE}
}

@article{varon2018quantifying,
  title={Quantifying methane point sources from fine-scale satellite observations of atmospheric methane plumes},
  author={Varon, Daniel J and Jacob, Daniel J and McKeever, Jason and Jervis, Dylan and Durak, Berke OA and Xia, Yan and Huang, Yi},
  journal={Atmospheric Measurement Techniques},
  volume={11},
  number={10},
  pages={5673--5686},
  year={2018},
  publisher={Copernicus GmbH}
}

@inproceedings{ronneberger2015u,
  title={U-net: Convolutional networks for biomedical image segmentation},
  author={Ronneberger, Olaf and Fischer, Philipp and Brox, Thomas},
  booktitle={Medical Image Computing and Computer-Assisted Intervention--MICCAI 2015: 18th International Conference, Munich, Germany, October 5-9, 2015, Proceedings, Part III 18},
  pages={234--241},
  year={2015},
  organization={Springer}
}

@article{zang2021land,
  title={Land-use mapping for high-spatial resolution remote sensing image via deep learning: A review},
  author={Zang, Ning and Cao, Yun and Wang, Yuebin and Huang, Bo and Zhang, Liqiang and Mathiopoulos, P Takis},
  journal={IEEE Journal of Selected Topics in Applied Earth Observations and Remote Sensing},
  volume={14},
  pages={5372--5391},
  year={2021},
  publisher={IEEE}
}

@article{lin2013network,
  title={Network In Network},
  author={Lin, Min and Chen, Qiang and Yan, Shuicheng},
  journal={arXiv e-prints},
  pages={arXiv--1312},
  year={2013}
}

@article{terven2023comprehensive,
  title={A comprehensive review of YOLO: From YOLOv1 to YOLOv8 and beyond},
  author={Terven, Juan and Cordova-Esparza, Diana},
  journal={arXiv preprint arXiv:2304.00501},
  year={2023}
}

@article{wang2020solov2,
  title={Solov2: Dynamic and fast instance segmentation},
  author={Wang, Xinlong and Zhang, Rufeng and Kong, Tao and Li, Lei and Shen, Chunhua},
  journal={Advances in Neural information processing systems},
  volume={33},
  pages={17721--17732},
  year={2020}
}

@inproceedings{he2017mask,
  title={Mask r-cnn},
  author={He, Kaiming and Gkioxari, Georgia and Doll{\'a}r, Piotr and Girshick, Ross},
  booktitle={Proceedings of the IEEE international conference on computer vision},
  pages={2961--2969},
  year={2017}
}

@inproceedings{long2015fully,
  title={Fully convolutional networks for semantic segmentation},
  author={Long, Jonathan and Shelhamer, Evan and Darrell, Trevor},
  booktitle={Proceedings of the IEEE conference on computer vision and pattern recognition},
  pages={3431--3440},
  year={2015}
}

@article{chen2017deeplab,
  title={Deeplab: Semantic image segmentation with deep convolutional nets, atrous convolution, and fully connected crfs},
  author={Chen, Liang-Chieh and Papandreou, George and Kokkinos, Iasonas and Murphy, Kevin and Yuille, Alan L},
  journal={IEEE transactions on pattern analysis and machine intelligence},
  volume={40},
  number={4},
  pages={834--848},
  year={2017},
  publisher={IEEE}
}

@article{HE202476,
title = {UB-FineNet: Urban building fine-grained classification network for open-access satellite images},
journal = {ISPRS Journal of Photogrammetry and Remote Sensing},
volume = {217},
pages = {76-90},
year = {2024},
issn = {0924-2716},
doi = {https://doi.org/10.1016/j.isprsjprs.2024.08.008},
author = {Zhiyi He and Wei Yao and Jie Shao and Puzuo Wang},
}

@article{he2015spatial,
  title={Spatial pyramid pooling in deep convolutional networks for visual recognition},
  author={He, Kaiming and Zhang, Xiangyu and Ren, Shaoqing and Sun, Jian},
  journal={IEEE transactions on pattern analysis and machine intelligence},
  volume={37},
  number={9},
  pages={1904--1916},
  year={2015},
  publisher={IEEE}
}

@article{SANIMOHAMMED2022100024,
title = {Instance segmentation of standing dead trees in dense forest from aerial imagery using deep learning},
journal = {ISPRS Open Journal of Photogrammetry and Remote Sensing},
volume = {6},
pages = {100024},
year = {2022},
issn = {2667-3932},
author = {Abubakar Sani-Mohammed and Wei Yao and Marco Heurich}
}

@ARTICLE{9027926,
  author={Jiang, Shenlu and Yao, Wei and Wong, Man Sing and Li, Gen and Hong, Zhonghua and Kuc, Tae-Yong and Tong, Xiaohua},
  journal={IEEE Journal of Selected Topics in Applied Earth Observations and Remote Sensing}, 
  title={An Optimized Deep Neural Network Detecting Small and Narrow Rectangular Objects in Google Earth Images}, 
  year={2020},
  volume={13},
  number={},
  pages={1068-1081},
  doi={10.1109/JSTARS.2020.2975606}
}

@inproceedings{zhou2018d,
  title={D-LinkNet: LinkNet with pretrained encoder and dilated convolution for high resolution satellite imagery road extraction},
  author={Zhou, Lichen and Zhang, Chuang and Wu, Ming},
  booktitle={Proceedings of the IEEE conference on computer vision and pattern recognition workshops},
  pages={182--186},
  year={2018}
}

@inproceedings{he2016deep,
  title={Deep residual learning for image recognition},
  author={He, Kaiming and Zhang, Xiangyu and Ren, Shaoqing and Sun, Jian},
  booktitle={Proceedings of the IEEE conference on computer vision and pattern recognition},
  pages={770--778},
  year={2016}
}

@article{shafiq2022deep,
  title={Deep residual learning for image recognition: a survey},
  author={Shafiq, Muhammad and Gu, Zhaoquan},
  journal={Applied Sciences},
  volume={12},
  number={18},
  pages={8972},
  year={2022},
  publisher={MDPI}
}

@article{veit2016residual,
  title={Residual networks behave like ensembles of relatively shallow networks},
  author={Veit, Andreas and Wilber, Michael J and Belongie, Serge},
  journal={Advances in neural information processing systems},
  volume={29},
  year={2016}
}

@article{2016On,
  title={On the expressive power of deep neural networks},
  author={ Raghu, Maithra },
  year={2016},
}

@inproceedings{bruggemann2021exploring,
  title={Exploring relational context for multi-task dense prediction},
  author={Br{\"u}ggemann, David and Kanakis, Menelaos and Obukhov, Anton and Georgoulis, Stamatios and Van Gool, Luc},
  booktitle={Proceedings of the IEEE/CVF international conference on computer vision},
  pages={15869--15878},
  year={2021}
}

@inproceedings{liu2019end,
  title={End-to-end multi-task learning with attention},
  author={Liu, Shikun and Johns, Edward and Davison, Andrew J},
  booktitle={Proceedings of the IEEE/CVF conference on computer vision and pattern recognition},
  pages={1871--1880},
  year={2019}
}

@inproceedings{he2018hashing,
  title={Hashing as tie-aware learning to rank},
  author={He, Kun and Cakir, Fatih and Bargal, Sarah Adel and Sclaroff, Stan},
  booktitle={Proceedings of the IEEE Conference on Computer Vision and Pattern Recognition},
  pages={4023--4032},
  year={2018}
}

@inproceedings{10.1145/3429309.3429328,
author = {Shelton, Jacquelyn and Polewski, Przemyslaw and Yao, Wei},
title = {U-Net for Learning and Inference of Dense Representation of Multiple Air Pollutants from Satellite Imagery},
year = {2021},
isbn = {9781450388481},
publisher = {Association for Computing Machinery},
address = {New York, NY, USA},
booktitle = {Proceedings of the 10th International Conference on Climate Informatics},
pages = {128–133},
numpages = {6},
location = {virtual, United Kingdom},
series = {CI2020}
}

@article{JIANG202330,
title = {Space-to-speed architecture supporting acceleration on VHR image processing},
journal = {ISPRS Journal of Photogrammetry and Remote Sensing},
volume = {198},
pages = {30-44},
year = {2023},
issn = {0924-2716},
author = {Shenlu Jiang and Yuliya Tarabalka and Wei Yao and Zhonghua Hong and Guofu Feng}

}

@article{JIA202314,
title = {Joint learning of frequency and spatial domains for dense image prediction},
journal = {ISPRS Journal of Photogrammetry and Remote Sensing},
volume = {195},
pages = {14-28},
year = {2023},
issn = {0924-2716},
author = {Shaocheng Jia and Wei Yao}
}

@article{POLEWSKI2021297,
title = {Instance segmentation of fallen trees in aerial color infrared imagery using active multi-contour evolution with fully convolutional network-based intensity priors},
journal = {ISPRS Journal of Photogrammetry and Remote Sensing},
volume = {178},
pages = {297-313},
year = {2021},
author = {Przemyslaw Polewski and Jacquelyn Shelton and Wei Yao and Marco Heurich}
}
%\begin{IEEEbiographynophoto}{Shiliang Fu}

%\newpage

%\begin{IEEEbiographynophoto}{Wei Yao}


%\vfill


\end{document}